\documentclass[a4paper,conference]{IEEEtran}
\usepackage{cite}

\ifCLASSINFOpdf
   \usepackage[pdftex]{graphicx}
   \DeclareGraphicsExtensions{.pdf,.jpeg,.png}
\else
\fi
\usepackage{amsmath}

\usepackage{amsfonts}
\usepackage{url}

\newcommand{\madima}{MADiMa }

\usepackage{enumitem}
\usepackage[dvipsnames]{xcolor}

\begin{document}
%
\title{Partially Supervised Multi-Task Network for Single-View Dietary Assessment}

\author{\IEEEauthorblockN{Ya Lu,
Thomai Stathopoulou and
Stavroula Mougiakakou}
\IEEEauthorblockA{ARTORG Center for Biomedical Engineering Research, University of Bern\\
\{ya.lu, thomai.stathopoulou, stavroula.mougiakakou\}@artorg.unibe.ch}}

\maketitle

\begin{abstract}
Food volume estimation is an essential step in the pipeline of dietary assessment and demands the precise depth estimation of the food surface and table plane. Existing methods based on computer vision require either multi-image input or additional depth maps, reducing convenience of implementation and practical significance. Despite the recent advances in unsupervised depth estimation from a single image, the achieved performance in the case of large texture-less areas needs to be improved. In this paper, we propose a network architecture that jointly performs geometric understanding (\textit{i.e.}, depth prediction and 3D plane estimation) and semantic prediction on a single food image, enabling a robust and accurate food volume estimation regardless of the texture characteristics of the target plane. For the training of the network, only monocular videos with semantic ground truth are required, while the depth map and 3D plane ground truth are no longer needed. Experimental results on two separate food image databases demonstrate that our method performs robustly on texture-less scenarios and is superior to unsupervised networks and structure from motion based approaches, while it achieves comparable performance to fully-supervised methods.
\end{abstract}
\IEEEpeerreviewmaketitle

\section{Introduction}

Dietary assessment for calorie and macro-nutrient estimation has become indispensable for individuals willing to follow a healthy life style~\cite{importdiet}. The traditional approaches based on self-maintained dietary records~\cite{choestimation} and food frequency questionnaires~\cite{questionair} remain time consuming and/or prone to errors~\cite{choestimation}. On the other hand artificial intelligence (AI) techniques~\cite{personalizedfood, joachim3D, im2calories, cvpr2019_1, recipe1m, gocarb} open new possibilities to automatically perform dietary assessment by directly analyzing food images. This novel concept has motivated extensive studies over the last decade~\cite{multiview3, gocarb, personalizedfood, joachim3D, im2calories}, which have nowadays become primary solutions for monitoring the daily nutrient intake of individuals.

\begin{figure}[t]
\begin{center}
\includegraphics[width=2.5in]{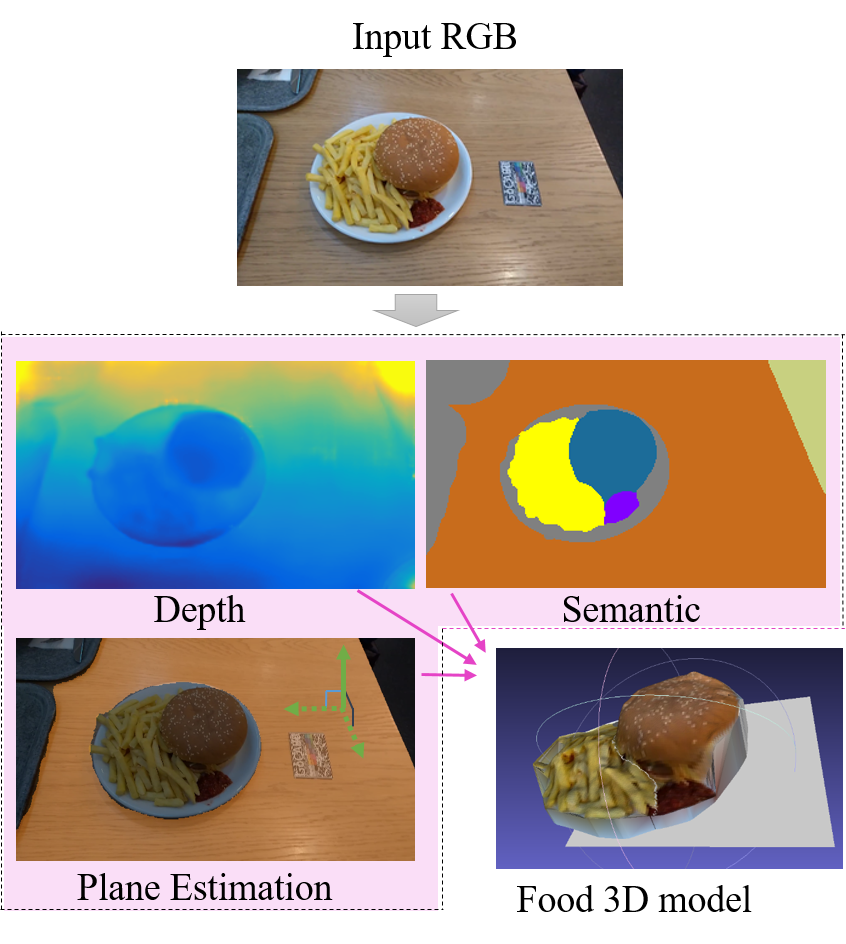}
\end{center}
   \caption{The input of the proposed system is a single RGB food image; the outputs are the predicted depth map, the semantic prediction map and the table plane. The food 3D model is built based on these three outputs.}
\label{fig:brief}
\end{figure}

Based on these AI-enabled dietary assessment approaches, the targeted nutrient information is typically inferred by firstly carrying out three image-processing steps~\cite{im2calories, gocarb}: i) food recognition; ii) food segmentation and iii) volume estimation, and then extracting the nutrient information from the relevant food composition databases (\textit{e.g.}, USDA database~\cite{USDA}) based on the results of steps i)-iii). The first two steps are usually treated by directly employing well established segmentation and recognition algorithms~\cite{grabpayMM, Food101, personalizedfood, mariosfoodclass}, which can both be implemented using one RGB-image input. However, the implementation of traditional food volume estimation techniques~\cite{voladd1, voladd2, multi3D}, such as Structure from Motion (SfM)~\cite{voladd2}, require more than one food images, and the performance is inevitably compromised by the almost always unknown information of the unseen part of the food (\textit{i.e.} the food bottom side, see Figure~\ref{fig:brief}). Although this limitation can be circumvented by estimating the table surface as a reference plane~\cite{im2calories, joachim3D, multiview3}, the texture-less quality of most tables impairs the ability to map points of interest between the input images, which further decreases the robustness and estimation accuracy. A more robust 3D food model reconstruction can be made by processing one RGB-image and the corresponding depth image captured from a depth sensor~\cite{point2volume, embc2019}, which can perform an accurate table plane estimation regardless of the texture of the table surface. However, the implementation of such approaches remains inconvenient as the depth sensor is not universally available to all end-users. As a simpler alternative, supervised Convolutional Neural Networks (CNNs) permit the use of a singular RGB-image as input~\cite{im2calories, madima2017}. These networks, however require depth images for their training, therefore a large and densely annotated training dataset, which is costly and time consuming.

In this paper, we propose a multi-task network architecture that jointly predicts the depth map, table plane and semantic map from only a single-view food RGB-image, with no need of a depth image in both training and testing processes, thus providing the ultimate convenience. Based on the outputs of said network, the food 3D model can be straightforwardly reconstructed using the Delaunay triangulation~\cite{geobook,joachim3D}, as shown in Figure~\ref{fig:brief}.

\begin{figure}[t]
\begin{center}
\includegraphics[width=1.0\linewidth]{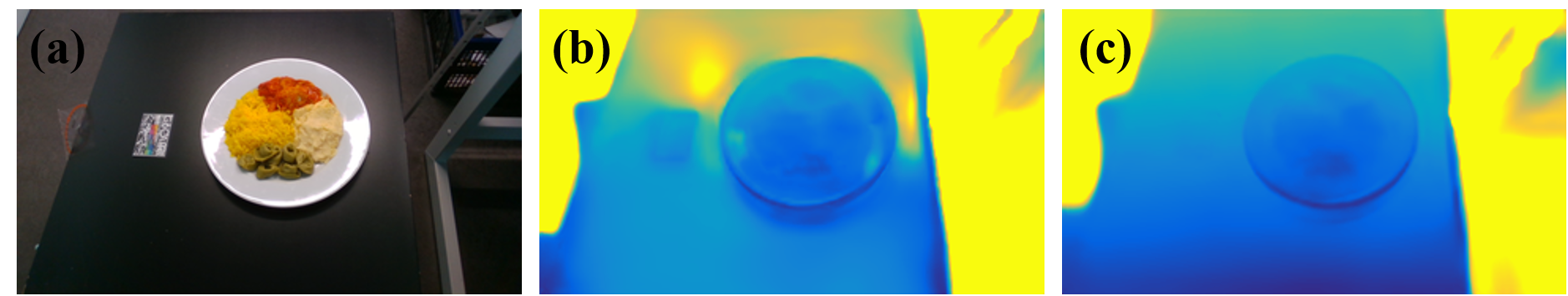}
\end{center}
   \caption{Single view depth prediction supervised by monocular videos: state-of-the-art~\cite{yin2018geonet} (b) and ours (c). (a) is the input RGB image. }
\label{fig:sample}
\end{figure}

In our proposed network, we use the self-supervised depth map prediction framework as the basic architecture~\cite{zhou2017unsupervised, yin2018geonet}, in which the monocular videos are the only supervision for the network training. In addition to this, two novel modules designed for the food scenarios, i.e. semantic prediction (supervised) and table plane estimation (unsupervised) modules, are integrated into the framework and jointly perform with the depth prediction module. This design does not only merge the entire dietary assessment pipeline within a single network, but also boosts the performance of depth prediction, especially in texture-less scenarios (see Figure \ref{fig:sample}). The network we propose can be trained using an end-to-end learning process.

The performance of the proposed network is evaluated on two databases containing food videos and semantic annotations: 1) the \madima database~\cite{madima2017} created in a laboratory setup and 2) a new database named ``Canteen database'', which is collected from three canteens. During the training data preparation, we propagate the semantic annotations for all the video frames using a small number of manually annotated frames, which significantly reduces the human effort.

\noindent
The contribution of this paper is three-fold:
\begin{enumerate}[label=\roman*]
    \item The 3D food model is reconstructed based on a single RGB-image input, completely avoiding the use of food depth image in either training or testing processes, thus for the first time to the best of our knowledge, fully realizing the conception of AI-based dietary assessment.
    \item The depth map, semantic map and table plane are jointly predicted by a single partially supervised network, enabling the full pipeline dietary assessment. With this design, the proposed network outperforms the SfM and unsupervised techniques, while exhibiting a comparable performance to fully supervised approaches.
    \item A new dietary assessment database with food videos captured in canteens is introduced and benchmarked.
\end{enumerate}

\section{Related work}

\subsection{Food image processing}

The majority of food image processing techniques focus on food recognition~\cite{Food101,recipe1m,mariosfoodclass} and segmentation~\cite{grabpayMM, UNIMIBdatabase}. 
With the fast development of CNNs, the performance of these two steps~\cite{grabpayMM, cnnfoodrecog, recipe1m} can nowadays significantly outperform the traditional approaches based on the usage of hand-crafted features~\cite{mariosfoodclass, UNIMIBdatabase, Food101}.

However, research rarely aims to address the food volume estimation problem. At the early stage, the multi-view SfM-based approaches are utilized to estimate the food volume~\cite{earlyMultiVolume, multiview4, joachim3D}. However, such approaches poorly perform in texture-less scenarios, and several additional constrains are imposed for the table plane estimation. Recently, the depth map produced by a depth sensor has been utilized for food volume estimation~\cite{point2volume, embc2019, madima2017}. Although such methods have achieved high accuracy in most scenarios, they require an additional depth image as input and may not perform well for black and reflective objects due to the intrinsic limitation of the depth sensors~\cite{madima2017}. On the other hand, the studies reported in~\cite{im2calories, madima2018, madima2017} predict the depth map from a single food image using supervised CNNs for 3D food model building~\cite{im2calories, madima2017} or the direct food volume regression~\cite{madima2018}. The performance of such methods greatly depends on the available densely annotated training databases, which are however costly and inconvenient for real applications.

There are also several studies investigating the full pipeline dietary assessment, such as GoCARB~\cite{gocarb} and im2calories~\cite{im2calories}. These systems, however, address the three steps of the pipeline as independent stages, while the contextual information among the tasks has not yet been fully exploited.

\begin{figure*}[htbp]
\begin{center}
\includegraphics[width=1.0\linewidth]{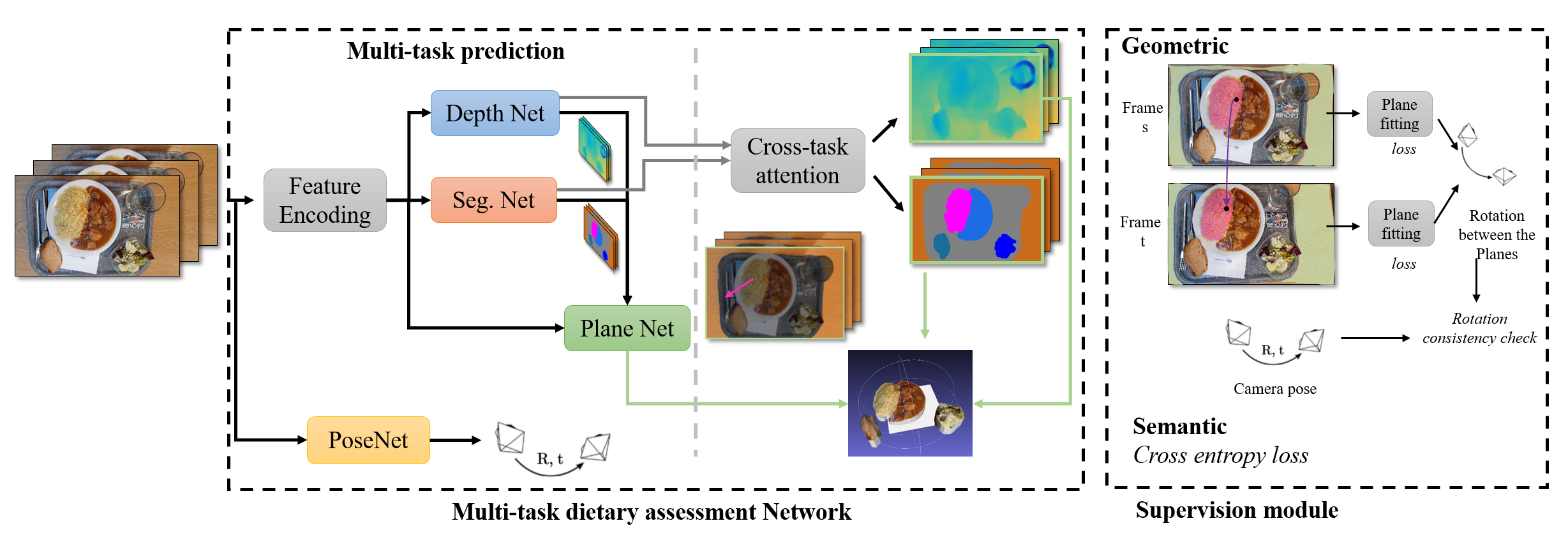}
\end{center}
   \caption{Overview of the proposed framework.}
\label{fig:overview}
\end{figure*}

\subsection{Geometry understanding from single color image}
\paragraph*{Depth estimation} Traditionally, the depth estimation from a single RGB image relies on hand-crafted features and strong assumptions~\cite{earlydepthSingle1, earlydepthSingle2}, which are difficult to generalize. \cite{eigenDepth, eigenImprove1} consider the task as a simple pixel regression problem and train the depth prediction model using end-to-end approaches. However, such methods require depth maps for training, which is costly to get for real scenario applications.
Recently, the unsupervised depth prediction from a single RGB image has been investigated. The depth image is not needed, and the networks are either supervised by a stereo image pair~\cite{unstereo1, mono_iccv2019} or monocular videos~\cite{zhou2017unsupervised, yin2018geonet, Meng2019signet, mono_iccv2019}. The latter is the most relevant to our work due to the simplicity of the training data collection. 
However, since the pixel-level view synthesis loss acts as the fundamental guideline during the network optimization, a poor performance is possible on the scenario with large texture-less area. Although \cite{Meng2019signet} and \cite{googleDepth} apply the semantic maps as additional inputs to the network, which may partially alleviate the issue, the problem still exists when the different texture-less areas have identical semantic labels.

\paragraph*{Plane estimation} In earlier studies, geometric cues, such as parallel lines and vanishing points were used for the estimation of 3D plane orientation~\cite{earlyPlane1,earlyPlane2}, which rely on the regular structures existing in the scene, thus limiting the generalization of these methods. To this end, machine learning-based approaches~\cite{learnPlane1,learnPlane2} are proposed and are used to overcome the generalization issue. However, all the aforementioned methods require either the plane parameters or additional depth maps as ground truth for the network training. To the best of our knowledge, there is no approach that can predict the 3D plane in an unsupervised way.

\section{Methodology}
In this section, a brief review of the monocular video supervised depth prediction framework is provided~\cite{zhou2017unsupervised,yin2018geonet}.
Then, a description of the proposed network architecture is given. Finally, the newly designed loss functions applied for the network optimization are introduced.

\subsection{Monocular unsupervised geometric understanding}
\label{sec:prob}
The inputs of the algorithm include $n$ consecutive image frames $I={\{I_i\}}_{i=1}^n, I_i\in\mathbb{R}^{H{\times}W{\times}3}$ from a video, along with the camera's intrinsic matrix $K\in\mathbb{R}^{{3}{\times}{3}}$. One of the frames (usually the center frame) is selected as "target frame": $I_t$ , while the rest are treated as the "source frames": $I_s$. In the algorithm, an encoding-decoding style CNN is built to predict the dense depth map $D\in\mathbb{R}^{H{\times}W}$ from each frame, while an ego-motion prediction network is jointly trained to estimate the camera transformation matrix $T_{t{\rightarrow}s}$ (composed by a rotation $R\in\mathbb{R}^{{3}{\times}{3}}$ and a translation $t\in\mathbb{R}^3$) between the target frame and source frames. Then, the rigid flow between the target and one source frame can be retrieved using:
\begin{equation}
\label{eq:rigidflow}
f_{t{\rightarrow}s}(p_t) = KT_{t{\rightarrow}s}D_{t}(p_t)K^{-1}p_t-p_t
\end{equation}
where $p_t$ indexes the pixel position of the target frame in homogeneous coordinates. 
During training, \eqref{eq:rigidflow} is used as the warping function to synthesize the target frame ($I_s^w$) from the source frame $I_s$, using the differentiable bilinear sampling mechanism~\cite{inversewarp}. The pixel value similarity (\textit{i.e.}, view synthesis loss, $\mathcal{L}_{vs}$) between the warped source frames $I_s^w$ and the target frames $I_t$ is utilized to guide the network optimization~\cite{yin2018geonet,zhou2017unsupervised}. Furthermore, to filter out erroneous predictions and preserve sharp details, the edge-aware smoothness loss ($\mathcal{L}_{s}$)~\cite{yin2018geonet,Meng2019signet} is also applied. 


\subsection{Network architecture}
The left side of Figure~\ref{fig:overview} depicts the overview of the proposed network architecture, which contains two main components: i) a multi-task prediction component that simultaneously predicts the depth map, semantic map and table plane from the single input image, and estimates the camera pose between the consecutive frames; ii) a cross task attention mechanism applied for the refining of the depth and semantic predictions.

\subsubsection{Multi-task prediction}
\label{sec:multi}
In this work, the multi-task prediction component encompasses 4 branches: \textit{Depth Net}, \textit{Seg. Net}, \textit{Plane Net} and \textit{Pose Net}. The first 3 branches share the same feature encoding module, which could basically employ any network structures, such as widely used VGG~\cite{vgg} and ResNet~\cite{resnet}. Here, we utilize the ResNet50~\cite{resnet} following the same way as~\cite{yin2018geonet}.

\begin{figure}[t]
\begin{center}
\includegraphics[width=1.0\linewidth]{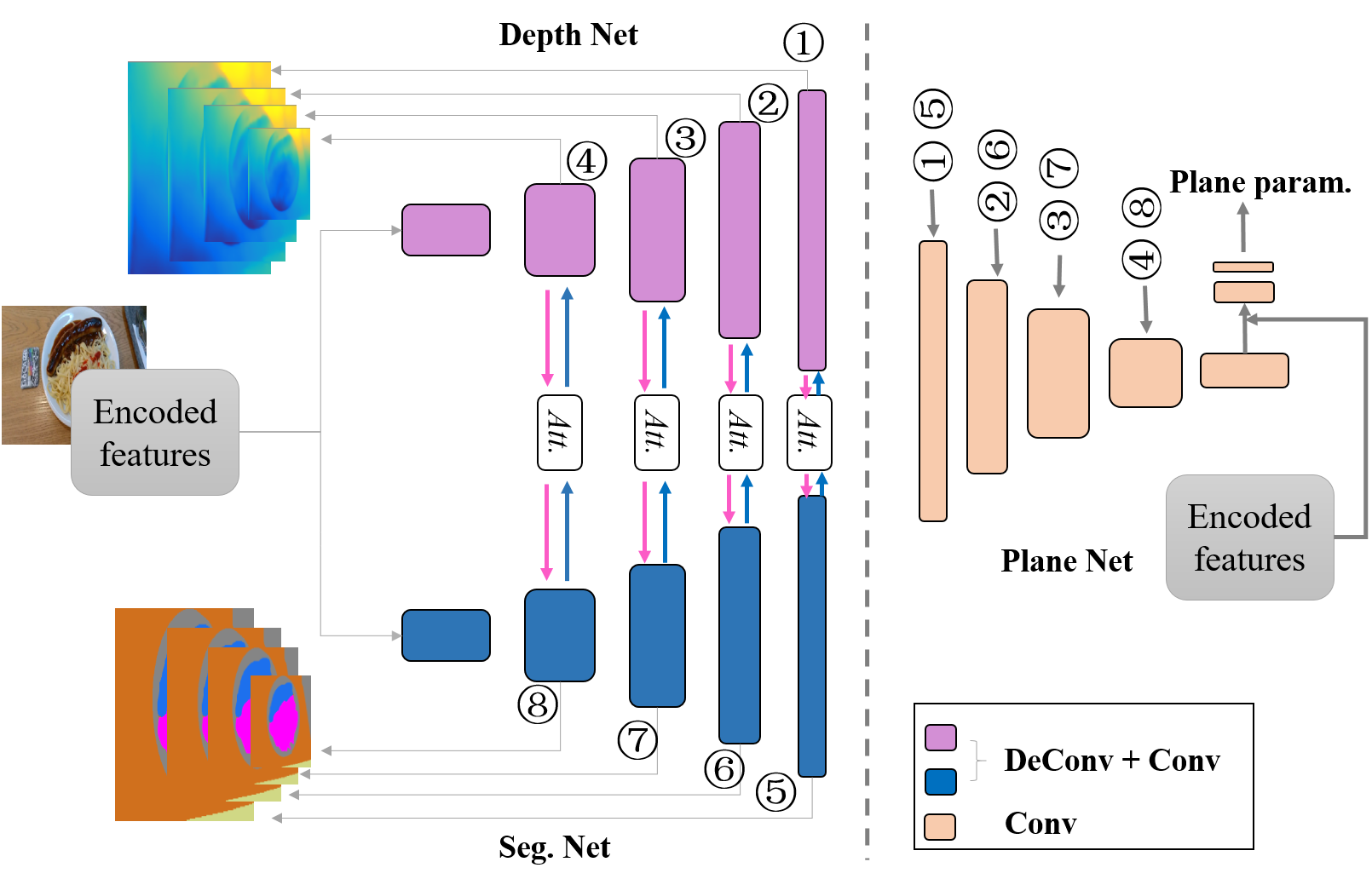}
\end{center}
   \caption{Network architecture of \textit{Depth Net}, \textit{Seg. Net} (left) and \textit{Plane Net} (right). The width and height of each block indicate the channel and the spatial dimension of the feature maps, respectively. Each reduction/increase in size indicates a change by a factor of 2. The first deconv/conv layer of \textit{Depth Net} and \textit{Seg. Net} has 512 channels, while that of  the \textit{Plane Net} has 32 channels. The filter size is 3 for all the conv/decov layers.   }
\label{fig:arc}
\end{figure}

The \textit{Depth Net} and \textit{Seg. Net} modules act as the decoder of the features encoded from ResNet50 to get the predictions of the depth and semantic maps, respectively. Both modules are made by convolutional / deconvolutional layers and skip connections. The detailed network parameters can be found in Figure~\ref{fig:arc}.

A normal way to represent a 3D plane is $(\tilde{\mathbf{n}}, d)$, where $\tilde{\mathbf{n}}\in\mathbb{R}^3$ is the normal vector of the plane, $d$ is the distance between the plane and camera center. Due to the fact that the camera center point should not pass the table plane in any real scenario, the table plane can be then simplified as $\mathbf{n}\in\mathbb{R}^3$, where $\mathbf{n}=\tilde{\mathbf{n}} /d$. Thus, the 3D plane in our network is represented as a 3-element vector.\par 
The input of the \textit{Plane Net} employs the predictions from \textit{Depth Net} and \textit{Seg. Net}, together with the output features from ResNet50. In the \textit{Plane Net}, the predictions from \textit{Depth Net} and \textit{Seg. Net} with 4 scales are consecutively concatenated and downscaled by 4 convolutional layers, which are then concatenated with the feature maps from ResNet50 after a convolutional layer (See the right part of Figure \ref{fig:arc} for detail). Finally, the parameters of the table plane are predicted after two convolutional layers and a global average layer.

The \textit{Pose Net} predicts the camera pose between the consecutive frames, and is adopted from the same architecture as in~\cite{zhou2017unsupervised, yin2018geonet}.

All the convolutional and deconvolutional layers in this multi-task prediction component are followed by the batch normalization and ReLU activation layers, except for the prediction layers\footnote{The softmax is applied on the \textit{Seg. Net} prediction layer, while $1/(\alpha\ast{sigmoid(x)}+\beta)$ with $\alpha=10$ and $\beta=0.5$ is performed for the \textit{Depth Net} prediction layer. No activation layer is applied on the prediction layers of \textit{Pose Net} and \textit{Plane Net}}.

\subsubsection{Cross-task attention}
In the current work, distinct loss functions are utilized for different learning tasks; however, there is evidence showing that in this case the performance of some tasks may be degraded~\cite{multiloss2} due to the non-ideal relative weight set for each task~\cite{multiloss1}. In addition, it is difficult and expensive to manually tune the parameters. 
Here we apply an attention mechanism on the food image depth and semantic prediction tasks, like in \cite{xu2018PAD-Net}, to allow the network to automatically select the useful information between the tasks during training. We formulate the attention mechanism using:

\begin{equation}
\label{eq:atten}
\begin{aligned}
&Depth_{final}\leftarrow{Depth}+\sigma(W_d^d{\cdot}Depth)\odot(W_s^d{\cdot}Seg) \\
&Seg_{final}\leftarrow{Seg}+\sigma(W_s^s{\cdot}Seg)\odot(W_d^s{\cdot}Depth)
\end{aligned}
\end{equation}
where $Depth$ and $Seg$ are the outputs of $Depth Net$ and $Seg. Net$, respectively. The $\sigma$ indicates the sigmoid function, while the $W$ are the weights of the convolutional layers and $\odot$ indicates element-wise multiplication.

\subsection{Loss functions}
Apart from the pixel-level view synthesis loss and smooth loss that have been used in the state-of-the-art unsupervised geometric understanding framework~\cite{yin2018geonet,zhou2017unsupervised,Meng2019signet} (also mentioned in Section~\ref{sec:prob}), three additional loss functions are jointly designed and utilized for the network optimization.

\subsubsection{Semantic loss}
Two types of semantic losses are applied during the network training:

The first one is the semantic segmentation loss ($\mathcal{L}_{s_p}$), which is used for the optimization of the semantic label prediction. We apply the typical cross-entropy loss function for it. During training, both the intermediate results and the final segmentation results (\textit{i.e.}, $Seg$ and $Seg_{final}$ in \eqref{eq:atten}) are contributed for $\mathcal{L}_{s_p}$ computation.

The second is the semantic constraint loss, which is used to boost the performance of the unsupervised geometric prediction. Such loss encourages the corresponding pixels in consecutive frames retrieved by rigid warping to be allocated with the same semantic labels (Figure~\ref{fig:overview}). The loss function is formulated as:
\begin{equation}
\label{eq:seg_loss}
\mathcal{L}_{s_c}=\|S_{s}^{w}-S_{t}\|_2
\end{equation}
where $S_t$ and $S_{s}^{w}$ are the one-hot semantic map of the target frame and the warped source frame, respectively. The $\mathcal{L}_{s_c}$ can improve the performance of geometric prediction in the case that two nearby distinct food items have similar pixel and texture appearance. 
It should be noted that the semantic maps we used in \eqref{eq:seg_loss} are from the annotated ground truth, thus the efficiency and generalization ability during the network optimization can be ensured. 

\subsubsection{Plane fitting loss}
From the predicted depth and semantic maps, the 3D points that belong to the table plane can be retrieved using: $P=D(p_{tab.})K^{-1}p_{tab.}$, where $p_{tab.}$ indexes the pixels that belong to the table in homogeneous coordinates and $D(p_{tab.})$ indicates the depth value of $p_{tab.}$. 
Well estimated plane parameters should ensure a small distance between $P$ and the plane, thus the plane fitting loss is defined as: 
\begin{equation}
\label{eq:plane_loss}
\mathcal{L}_{p}=\frac{1}{N_{tab.}}\sum_{P\in{tab.}} \vert{{\mathbf{n}}^TP-1}\vert
\end{equation}
where $\mathbf{n}$ is the 3D plane parameters (more details can be found in Section~\ref{sec:multi}), $\vert{{\mathbf{n}}^TP-1}\vert$ computes the absolute distance from $P$ to the plane surface. The plane fitting loss in \eqref{eq:plane_loss} can not only act as the guidance for the optimization of the table plane estimation, but also boost the performance of the depth prediction by encouraging the pixels on the table locating to the same plane in 3D space. 

\subsubsection{Consistency loss}
Maximizing the consistency between the table plane rotation predicted by \textit{Plane Net} and the camera pose rotation predicted by \textit{Pose Net} can improve the accuracy of the table plane estimation. To do so, a consistency loss is introduced and formulated as: 
\begin{equation}
\label{eq: consist}
\mathcal{L}_{c} = \parallel{\Delta{q_{s\rightarrow{t}}^p}}-\Delta{q_{t\rightarrow{s}}^c}\parallel_2
\end{equation}
where the $\Delta{q_{s\rightarrow{t}}^p}$ is the quaternion rotation vector from the table plane of the source image to that of the target image, while the $\Delta{q_{t\rightarrow{s}}^c}$ is the camera rotation from the target image to source image.


\section{Databases}
\label{sec:db}

Two databases are used to demonstrate the performance of the proposed approach in laboratory setup and real canteen scenario.

\begin{figure}[t]
\begin{center}
\includegraphics[width=1.0\linewidth]{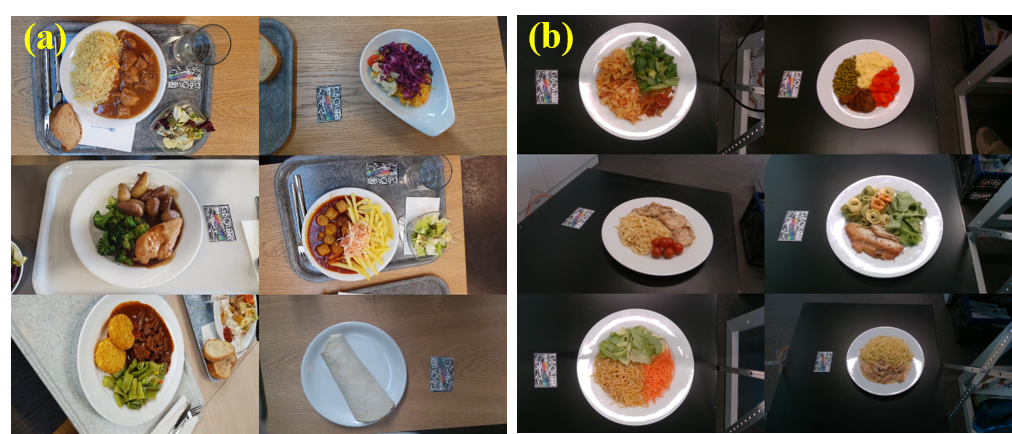}
\end{center}
   \caption{Some meal samples contained in Canteen database (a) and \madima database (b).}
\label{fig:DB}
\end{figure}

\textbf{\madima database} The database is created in a laboratory setup and contains 80 central-European style meals, including 234 food items of known volume. The meals are placed on a black, texture-less and reflective table during the image capturing, aiming at simulating the least appropriate setup for geometric understanding (Figure~\ref{fig:DB}). For each meal, an RGBD video sequence with 200 frames is collected. We split the dataset into 60 meals for training, as in~\cite{madima2017} and the remaining 20 for testing. Thus, in total, there are 12$k$ images for training and 4$k$ images for testing. Since the semantic annotation is not provided for the video sequence in the original database, we manually annotate 5 frames of each meal, while the remaining 195 frames are propagated automatically using the pre-trained PSPNet~\cite{pspnet}. This way the annotation effort is reduced by 95\%. The annotated classes include the plate area, the table area and 6 food categories as defined in~\cite{madima2018}.

\textbf{Canteen database} The database is collected from three different canteens, containing 80 meals for training, and 10 meals for testing. For each training meal, a short video with $\sim$200 RGB frames is captured using normal smart phones, corresponding to $\sim$16$k$ food images in total. The semantic segmentation maps are annotated and include the table area, plate area and 6 hyper food categories (\textit{i.e.}, vegetable, meat, potato, grain, bread, sauce) covering most of the canteen meals. Similar to the \madima databse, the PSPNet\cite{pspnet} is used to propagate the annotations.

For each meal in the testing set, 6 RGBD image pairs are captured at different angles of view using the Intel depth camera, for the use of depth prediction evaluation. In addition, another 6 RGB images with the associated gravity data are collected using smartphones, for the evaluation of the table plane estimation. The table plane orientation is annotated using the reversed gravity vector of the smartphone, based on the assumption that the table plane is horizontal in the world coordinates. The semantic maps are also annotated on these 6 RGB images for the evaluation of food semantic segmentation. In addition, a video with $\sim$200 frames is collected for the food volume annotation using the same way as in~\cite{madima2017}. 
 
\section{Experimental results}
\subsection{Experimental setup}
The proposed network is trained from scratch using the TensorFlow framework. During training, the image sequences are resized to a resolution of 180$\times$320, while the length of the sequence is fixed at 3. Similar with the hyper parameters setup in~\cite{yin2018geonet, zhou2017unsupervised}, we use the Adam optimizer with initial learning rate of $2e-4$ and the batch size is set to 4. The training data are augmented by scaling, random cropping and color jittering. We optimize the network in two different stages. At the first stage, the weights of \textit{Plane Net} are fixed and we optimize the \textit{Depth Net}, \textit{Seg. Net} and \textit{Pose Net} for 100$k$ iterations. After that, another 200$k$ iterations are performed for the whole network optimization.
\begin{figure*}[t]
\begin{center}
\includegraphics[width=1.0\linewidth]{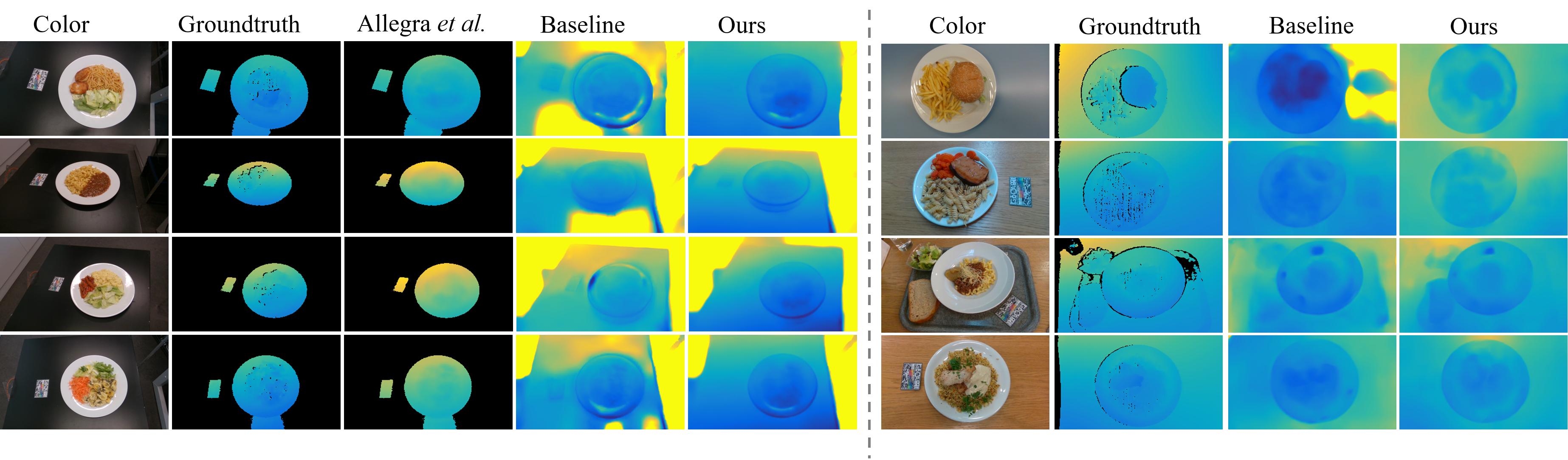}
\end{center}
   \caption{Visualization results on \madima (left) and Canteen (right) database. 
   The method used in ``Allegra \textit{et al.}"~\cite{madima2017} is based on the \textbf{supervised} learning.
    We only display the pixels that has valid value in ground truth, since ``Allegra \textit{et al.}''~\cite{madima2017}  can not provide reasonable prediction on the rest pixels. The baseline is implemented with the \textbf{unsupervised} approach in~\cite{yin2018geonet}.}
\label{fig:madimaR}
\end{figure*}

\subsection{Geometric prediction}
\label{sec:gp}
In this section, the performance of the geometric prediction including the depth prediction and table plane estimation are evaluated.
\begin{table*}
\begin{center}
\caption{Comparison results of depth estimation. ``M'' and ``C'' indicate \madima and Canteen database, respectively. The \textbf{bold} indicates the best performance with unsupervised approach, while the ``\underline{   }'' is the best performance of supervised method. }
\begin{tabular}{ l|c|c|c c c c|c c c }
\hline
              &      &                   &\multicolumn{4}{|c|}{Error metrics}&\multicolumn{3}{c}{Accuracy metrics}\\
Method & DB & Supervision & Abs. Rel. & Sq. Rel. & RMSE & RMSE log & $\delta<1.05$ & $\delta<1.05^2$& $\delta<1.05^3$\\

\hline\hline

Allegra \textit{et al.}~\cite{madima2017} & M & Depth & 0.017 & 0.279 & 11.63 & 0.023 &0.977&\underline{0.999}&\underline{1.0} \\
Lu \textit{et al.}~\cite{madima2018} & M & Depth& \underline{0.013} &\underline{0.181} & \underline{9.27} & \underline{0.018} &\underline{0.988}&\underline{0.999}&\underline{1.0}\\
\hline
GeoNet\cite{yin2018geonet} & M & Mono &0.028 & 1.719 &26.55&0.046&0.885&0.955&0.974\\
Monodepth2\cite{mono_iccv2019} & M & Mono &0.027 & 0.647 & 17.36 &  0.032 & 0.863 & 0.984 & \textbf{0.998}\\
Ours     & M & Mono &\textbf{0.022}& \textbf{0.488} &\textbf{14.86}&\textbf{0.029}&\textbf{0.907}&\textbf{0.989}&0.996\\
\hline\hline
GeoNet\cite{yin2018geonet}& C & Mono &0.080 & 4.160 & 29.90 &  0.097 &    0.434 &    0.721  &    0.873\\
Monodepth2\cite{mono_iccv2019} & C & Mono & 0.063 & 7.617 & 30.01 & 0.086 & 0.527 & 0.836 & 0.947\\
Ours& C & Mono &\textbf{0.056} & \textbf{1.536} &\textbf{20.53}& \textbf{0.070}&  \textbf{ 0.535} &    \textbf{0.834} &    \textbf{0.951}\\
\hline
\end{tabular}
\end{center}

\label{tab:ori}
\end{table*}

\subsubsection{Depth prediction}
We evaluate the performance of the depth prediction on both \madima and Canteen databases using the metrics introduced in~\cite{eigenDepth}. For the \madima database, the evaluation is conducted on the 4$k$ food images from the 20 meals. It should be noted that, since the ground truth value on the black table surface cannot be stably provided, we only evaluate the depth values inside the plate following the same procedures as~\cite{madima2017, madima2018}. In the Canteen database, all the pixels with valid ground truth values are considered in the evaluation. The system was tested using the 10 testing meals, each meal containing 6 images, thus the system was tested on 60 images. During the evaluation, we multiply the predicted depth maps by a scaling factor to match the median of the ground truth following the same way as~\cite{zhou2017unsupervised, yin2018geonet}. Depth is measured in millimeters ($mm$).

The performance of our approach is compared to the baseline framework~\cite{yin2018geonet} and the latest state-of-the-art~\cite{mono_iccv2019} on both databases. To achieve this, the model in~\cite{yin2018geonet,mono_iccv2019} is trained using our databases, while the code used in the experiments is from the official release of~\cite{yin2018geonet,mono_iccv2019}. 
In addition to this, we proceeded to a comparison with previously introduced supervised methods~\cite{madima2017, madima2018} on the \madima database.

Figure \ref{fig:madimaR} provides the visual comparisons between our results and the aforementioned state-of-the-art methods (\cite{yin2018geonet, madima2017}) on both databases. It can be observed that our method always performs better than the unsupervised baseline~\cite{yin2018geonet} on the texture-less areas and is able to capture more details on food area. It can also be noted that our approach can predict table depth information in all cases, while the supervised method~\cite{madima2017} is inherently ill-suited for the black table surface due to lack of good quality training data. The quantitative comparisons are provided in Table~\ref{tab:ori}, showing that the proposed approach outperforms the unsupervised state-of-the-arts~\cite{yin2018geonet, mono_iccv2019} for all evaluation metrics on both databases, while presenting slightly lower performance than the supervised approaches~\cite{madima2017, madima2018} on \madima database. However, it must be mentioned that the performance of the supervised approaches~\cite{madima2017, madima2018} highly rely on the objective color, thus exhibiting great lack of robustness. Furthermore, the evaluation is conducted only for the pixels inside the plate due to the inherent character of the database, while the pixels outside the plate are discarded. To thoroughly evaluate the performance gains from each newly proposed module, ablation studies are conducted. Table~\ref{tab:aba} shows the comparison results during the study. The results from the table demonstrate that each of the newly proposed module improves the performance of our method, while the introduction of \textit{Plane Net} mostly contributes to the algorithm performance on both databases.

\begin{table*}
\begin{center}
\caption{Depth prediction performance gains due to different modules. ``S." indicates \textit{Seg. Net}; ``A." is the Cross-task Attention; ``P." indicates \textit{Plane Net}; ``C." means the usage of the consistency loss in eq. \eqref{eq: consist}. `` $\surd$ " indicates the corresponding module is included in the ablation study. }
\begin{tabular}{c c c c|c|c c c c|c c c}
\hline
    &   &     &    &         &\multicolumn{4}{|c|}{Error metrics}&\multicolumn{3}{c}{Accuracy metrics}\\
S. & A. & P. & C. & DB  & Abs. Rel. & Sq. Rel. & RMSE & RMSE log & $\delta<1.05$ & $\delta<1.05^2$& $\delta<1.05^3$\\

\hline\hline
 & & & & M  &0.028 & 1.719 &26.55&0.046&0.885&0.955&0.974\\
$\surd$& & & & M  &0.027 &1.132 &23.05&0.042&0.872&0.953&0.980\\
$\surd$& $\surd$ & & & M  &0.028 & 1.054 &21.61&0.040&0.845&0.952&0.985\\
$\surd$& $\surd$&$\surd$ & & M  &0.023 & 0.635 &15.64&0.033&0.893&\textbf{0.990}&0.992\\
$\surd$& $\surd$ & $\surd$& $\surd$& M  &\textbf{0.022} &\textbf{0.488} &\textbf{14.86}&\textbf{0.029}&\textbf{0.907}&0.989&\textbf{0.996}\\
\hline\hline
& & & & C &0.080 & 4.160 & 29.90 &  0.097&    0.434 &    0.721  &    0.873\\
$\surd$& & & & C & 0.075&3.710&28.69&0.093&0.448&0.763&0.896 \\
$\surd$& $\surd$ & & &C&    0.073& 3.142 & 27.78 & 0.091 & 0.447 & 0.754 & 0.892 \\
$\surd$& $\surd$&$\surd$ & & C  &0.059 & 1.561 &\textbf{20.09}&  0.071&    0.531 &    \textbf{0.854} &    0.948\\
$\surd$&$\surd$ & $\surd$& $\surd$& C  &\textbf{0.056} & \textbf{1.536} &20.53&  \textbf{0.070}&    \textbf{0.535} &    0.835 &   \textbf{ 0.951}\\
\hline
\end{tabular}
\end{center}

\label{tab:aba}
\end{table*}

\begin{table}
\begin{center}
\caption{Comparison results of plane orientation prediction on the Canteen database.}
\begin{tabular}{l|c|c}
\hline
Method &  Img. Num. & OE  \\
\hline\hline
Dehais \textit{et al.}\cite{joachim3D} &  2 & 0.22 \\
Ours- \textit{w/o C.} & 1 & 0.16 \\
Ours &1 & \textbf{0.14} \\
\hline
\end{tabular}
\end{center}

\label{tab:Plane}
\end{table}

\subsubsection{Table plane orientation}
The evaluation is conducted on 60 food images from the Canteen database. We adopt the Orientation Error (OE)~\cite{planeEva} as the metric for table orientation evaluation, which is denoted as $acos(\tilde{\mathbf{n}}^{\mathrm{T}}\tilde{\mathbf{n}}^*)$, where $\tilde{\mathbf{n}}$ and ${\tilde{\mathbf{n}}^*}$ are the predicted and ground truth table plane orientations, respectively.

The experimental result obtained by our approach is compared with the two-view SfM-based approach~\cite{joachim3D, multiview3, madima2017}, which has been widely used for food volume estimation. The latter constrains the plate shape as ellipse, and the auxiliary view used in the experiments is randomly chosen from the images belonging to the same meal. The comparison quantified by the OE is shown in Table \ref{tab:Plane}, demonstrating the advancement of our approach. The result of ablation study of the consistency loss is also given, where the ``Ours- \textit{no C.}" indicates that the consistency loss is not used during the model training. It is evident that consistency loss function dose contribute to improve the performance.

\subsection{Dietary assessment}

\begin{figure}[t]
\begin{center}
\includegraphics[width=1.0\linewidth]{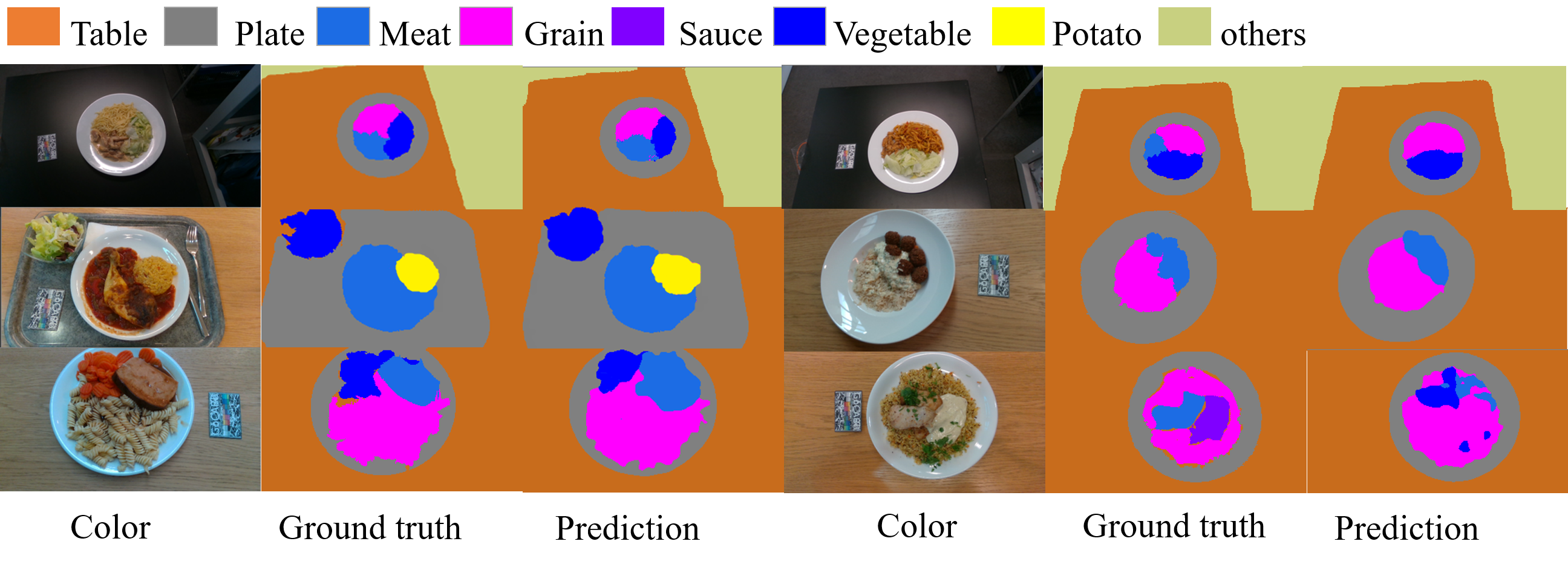}
\end{center}
   \caption{Examples of semantic segmentation results.}
\label{fig:seg}
\end{figure}

\subsubsection{Food image semantic segmentation}
The performance of the semantic segmentation is evaluated on 4$k$ testing images from the \madima database and 60 testing images from the Canteen database, respectively and is expressed by the standard mean Intersection over Union (mIoU) and pixel accuracy metrics shown in Table~\ref{tab:seg}. More than 90\% pixel accuracy is obtained on both databases, validating the good performing functionality of the specially designed \textit{Seg. Net}. An ablation study for the proposed cross-task attention module is also conducted and results are provided in Table~\ref{tab:seg}, where "Ours-\textit{no A.}" denotes that the cross-task attention module is removed during network training. It is clearly demonstrated that this module plays a positive role for the whole semantic estimation. Furthermore, it is worth mentioning that, some of the estimation errors indicated in Table~\ref{tab:seg} are caused by the overlap or similar appearance among the food items, as exemplified in Figure~\ref{fig:seg}. These errors are intrinsic in the food segmentation scenario and cannot be efficiently avoided by any of the known techniques.

\begin{table}
\begin{center}
\caption{Comparison results of food segmentation. ``M" and ``C" indicate the \madima and Canteen database, respectively.}
\begin{tabular}{l|c|c|c}
\hline
Method  & DB & mIoU & Pix. Acc\\

\hline\hline
Ours-\textit{w/o A.} &M& 68.4\% &95.9\%\\
Ours&M&\textbf{70.7\%}&\textbf{96.4\%}\\
\hline
Ours-\textit{w/o A.} & C &64.5\%&90.4\%\\
Ours &C&\textbf{68.5\%}&\textbf{91.4\%}\\

\hline
\end{tabular}
\end{center}

\label{tab:seg}
\end{table}

\subsubsection{Food volume estimation}
The evaluation is performed on 40 images from 20 different meals in the \madima database (\textit{i.e.}, the ``free set" defined in~\cite{madima2018}, which is a subset of the 4$k$ testing images in Section~\ref{sec:db}), and 60 images from 10 different meals in Canteen database, respectively. For a fair comparison with other methods, we provide the ground truth segmentation map combined with the predicted depth map and the table plane for the food 3D model building.

Using the Mean Absolute Percentage Error (MAPE) as the evaluation metric, Table~\ref{tab:volume} provides the quantitative volume estimation results on both databases, in which the results of \cite{joachim3D} and \cite{madima2018} on the \madima database are taken from~\cite{madima2018}. It can be seen that, our approach significantly outperforms the two-view SfM-based approach in \cite{joachim3D} on the \madima database and exhibits a good performance ($\sim$20\%) on the newly proposed Canteen database. It can also be observed that the proposed approach slightly underperforms the fully supervised approach~\cite{madima2018} on the \madima database; however, our technique does not require densely annotated ground truth data for training, and thus is capable of matching the vast diversity of real world applications.

\begin{table}
\begin{center}
\caption{Comparison results of food volume estimation. ``M" and ``C" indicate the \madima and Canteen database, respectively.}
\begin{tabular}{l|c|p{1cm}|c|c}
\hline
Method & Supervision & Img. Num. & DB & MAPE \\
\hline\hline
Lu \textit{et al.}\cite{madima2018} &Depth+Vol.& 1 & M &\underline{19.1\%} \\
Dehais \textit{et al.}~\cite{joachim3D}&Mono views&2 & M & 36.1\% \\
Ours & Mono &1 &M & \textbf{25.2\%}\\
\hline
\hline
Ours &Mono&1 &C & 20.3\%\\
\hline
\end{tabular}
\end{center}

\label{tab:volume}
\end{table}

\section{Conclusion}
In this paper, we propose a partially supervised network architecture that jointly predicts depth map, semantic segmentation map and 3D table plane from a single RGB food image, for the first time enabling the full-pipeline single-view dietary assessment to the best of our knowledge. The training procedure is only supervised by monocular videos with small number of semantic ground truth, with no need of both depth and 3D plane ground truth. Benefited from the specially designed semantic prediction module, 3D plane estimation module and loss functions, the proposed network significantly outperforms the SfM-based approach and the state-of-the-art unsupervised geometric prediction approach, while presenting a comparable performance with respect to the fully supervised approach.

\section*{Acknowledgment}
This research was funded in part by the SV Stiftung.



%

\bibliographystyle{IEEEtran}
\bibliography{IEEEabrv,egbib}




\end{document}